\documentclass[conference]{IEEEtran}
\IEEEoverridecommandlockouts
\usepackage{cite}
\usepackage{amsmath,amssymb,amsfonts}
\usepackage{algorithm}
\usepackage[french,vlined,ruled,algo2e]{algorithm2e}
\usepackage{algorithmic}
\usepackage{graphicx}
\usepackage{tabularray}
\usepackage{textcomp}
\usepackage{xcolor}
\usepackage{multirow}
\def\BibTeX{{\rm B\kern-.05em{\sc i\kern-.025em b}\kern-.08em
    T\kern-.1667em\lower.7ex\hbox{E}\kern-.125emX}}

\newcommand\blfootnote[1]{%
  \begingroup
  \renewcommand\thefootnote{}\footnote{#1}%
  \addtocounter{footnote}{-1}%
  \endgroup
}
\begin{document}

\title{Multi-modal Multi-view Clustering based on Non-negative Matrix Factorization}

\author{\IEEEauthorblockN{Yasser KHALAFAOUI*}
\IEEEauthorblockA{\textit{R\&D department} \\
\textit{ALTECA}\\
Massy, France \\
mykhalafaoui@alteca.fr}
\and
\IEEEauthorblockN{Nistor GROZAVU}
\IEEEauthorblockA{\textit{ETIS - CNRS UMR 8051} \\
\textit{CY Cergy Paris University}\\
Cergy, France \\
nistor.grozavu@cyu.fr}
\and
\IEEEauthorblockN{Basarab MATEI}
\IEEEauthorblockA{\textit{LIPN - CNRS UMR 7030} \\
\textit{Sorbonne Paris Nord University}\\
Villetaneuse, France \\
matei@lipn.univ-paris13.fr}
\and
\IEEEauthorblockN{Laurent-Walter GOIX}
\IEEEauthorblockA{\textit{R\&D department} \\
\textit{ALTECA}\\
Lyon, France \\
lwgoix@alteca.fr}

}

\maketitle

\begin{abstract}
By combining related objects, unsupervised machine learning  techniques aim to reveal the underlying patterns in a data set.\par
Non-negative Matrix Factorization (NMF) is a data mining technique that splits data matrices by imposing restrictions on the elements' non-negativity into two matrices: one representing the data partitions and the other to represent the cluster prototypes of the data set. This method has attracted a lot of attention and is used in a wide range of applications, including text mining, clustering, language modeling, music transcription, and neuroscience (gene separation). The interpretation of the generated matrices is made simpler by the absence of negative values. In this article, we propose a study on multi-modal clustering algorithms and present a novel method called \textit{multi-modal multi-view non-negative matrix factorization}, in which we analyze the collaboration of several local NMF models. The experimental results show the value of the proposed approach, which was evaluated using a variety of data sets, and the obtained results are very promising compared to state of art methods.
\end{abstract}

\blfootnote{* Corresponding author}
\section{Introduction}
The development and everyday use of social media has led people to share their lives and express their opinions online. As a result, data (text, images, audio/speech, video, etc.) generated by social networks users is changing rapidly. As data collections become highly diversified \cite{b1} due to the emergence of multi-modal data sets, multi-view data sets (i.e. the same data sample described in various ways) and dispersed data, it is now critical to effectively extract inherent information from these multi-source data sets. Data Clustering is an approach to discover the intrinsic structures of a collection of items by grouping objects with similar features \cite{b2}.\par
Due to the increasing variety and volume of data sets, clustering algorithms struggle to achieve competitive results with high certainty. However, similar issues can be addressed more easily by combining several approaches to improve both the quality and reliability of the outputs.\par
NMF has received a lot of attention in recent years \cite{b3} \cite{b4} and has been used in a variety of domains including feature selection, dimensionality reduction, text mining and clustering \cite{b2}. Paatero (1994) \cite{b5} established the NMF method, an unsupervised clustering methodology in which a data matrix is factored into (usually) two matrices: a matrix of cluster prototypes and a matrix of data partitions, such that none of the matrices has any negative component. The exclusion of negative values makes it easier to interpret the constructed matrices.
Self-Organizing Map (SOM) is another clustering algorithm that involves artificial neural networks \cite{b16}. To achieve clustering, this method processes all of the data samples one at a time and maps the cluster centers to a two-dimensional space.
De Sa (2005) \cite{b8} presented a simple and effective spectral clustering approach and used it to analyse web page data with two views. The similarity matrix is used first to combine the features extracted of both views, and then the standard spectral clustering technique is used to perform clustering and produce the final clustering result.\par
In a multi-view setting, a data sample can describe the same item from different angles and in different ways \cite{b6}. Having different views complementing each other, multi-view clustering algorithms become important for information extraction. In the literature, we distinguish four categories:
\begin{itemize}
    \item \textit{Multi-view graph clustering}. These methods find a fusion graph (or network) across all views and then applies semi-automatic segmentation algorithms or other techniques (e.g., spectral clustering) to the fusion graph to produce the clustering result. Wang et al. (2017) \cite{b12} introduced a generative model that uses ensemble manifold regularization. In particular, they built a nearest neighbor graph for each view to encode the corresponding manifold information, and a multiple graph ensemble regularization framework was designed to learn the optimal intrinsic manifold. The PLSA-based multi-view topic model was then modified to include the manifold regularization term, producing a unified objective function. Zhang and Mao (2016) \cite{b13} used sparse weights for similarity graph generation with unreliable neighbors filter in order to identify accurate neighbors for multi-view clustering efficiently, by presenting every object as a weighted sum of its neighbors for each view.
    \item \textit{Multi-kernel learning}. This class of methods employs predefined kernels associated to different views, which are then combined either linearly or non-linearly to improve clustering performance \cite{b9}. Zhao et al. (2009) \cite{b10} introduced a multi-kernel clustering algorithm based on maximum margin clustering, which finds the best clusterings, the optimal kernels as well as the maximum margin hyperplane together at the same time. Du et al. (2015) \cite{b11} proposed a robust K-means (with $l_{2,1}$-norm) on kernel space and applied a multiple kernel K-means algorithm that can find simultaneously the optimal combination of multiple kernels, the best clustering labels and cluster membership.
    \item \textit{Multi-task multi-view clustering}. These methods assign one or more tasks to each view, transfer inter-task knowledge to one another, and exploit multi-task and multi-view relationships to improve clustering performance.Gu and Zhou (2009) \cite{b15} presented a cross-domain based multi-task clustering solution in which each view is assigned a task. This method aims to learn a subspace that allows knowledge transfer from one task to another. Xie et al. (2012) \cite{b14} presented a 3-factor NMF-based multi-task collaborative clustering method. The cost function was made of two parts: task-specific co-clustering and cross-task feature space regularization.
    \item \textit{Collaborative clustering algorithms}. This approach deals with multi-view data by adopting a co-training strategy. It bootstraps the clustering of different views by using the information extracted from one another. By applying this method iteratively, the clustering results of all views tend to converge, leading to the broadest consensus across all views. Bickel and Scheffer (2004) \cite{b7} introduced a k-means-based multi-view clustering algorithm and applied it to text clustering data with two conditionally independent views. Furthermore, Grozavu et al. (2022) \cite{b2} proposed a NMF based multi-view clustering. First, NMF is applied to each view independently, then a collaboration phase is added in order to find hidden structures and patterns, and allow the interaction between these different views. 
\end{itemize}
However, these approaches are not adapted for  the multi-view multi-modal aspect of the data sets (i.e. multi-source data sets where each source can have multiple views or representations). Taking into account the augmenting complexity and volume of data sets nowadays and the need for efficient information extraction algorithms, it is important to develop a solution that tackles this subject.\par
Our research sets out to propose a new method for multi-modal multi-view clustering by extending some multi-view solutions proposed in the literature. The algorithm first applies a NMF on each view locally; then a collaboration phase between different views within the same modality allows for the exchange of information and finally a second collaboration phase is introduced, where each of the other modalities contributes to the co-clustering.\par
The remainder of the paper is organized as follows: Section II  discusses the preliminary setting of the proposed approach and the formal definition of our solution. Section III proposes optimizations of the solution under different conditions. Finally, we assess the performances of the proposed algorithm through experimental results in Section Experiment. The paper ends with a conclusion and several future works.

\section{Problem Formalization}
\subsection{NMF Algorithm}
The traditional Nonnegative Matrix Factorization algorithm is proven to be equivalent to relaxed K-means clustering method \cite{b17}. Given a non-negative data matrix of $M$ features and $N$ objects, denoted as $X = (x_1, x_2,\cdots,x_N) \in \mathbb{R}_+^{M\times N}$, such that $x_n \in \mathbb{R}_+^{M \times 1}$ represents the $n^{th}$ object of $X$, the NMF algorithm gives a low rank approximation of $X$ using two non-negative matrices product $FG$, such that $F$ is the matrix of cluster prototypes and $G$ is the matrix of data partitions defined respectively as $F = (f_1, f_2,\cdots,f_k) \in \mathbb{R}_+^{M\times K}$, and $G = (g_1, g_2,\cdots,g_N) \in \mathbb{R}_+^{K\times N}$, with $K$ a parameter representing the number of components. Under a constrained optimization, the NMF cost function to minimize can be written as:
\begin{equation}\label{eq:nmf_cf}
\mathcal{L}(X,F,G)= \lVert X - FG \rVert^2
\end{equation}
with $\mathcal{L}$ representing the Frobenius norm of the matrix $X - FG$.

\subsection{Multi-modal Multi-view Setting}
In this section, we investigate the exchange of information between finite clustering results obtained using an NMF model and those obtained in a multi-modal multi-view context. NMF clustering algorithm is applied on each data set. We are interested in the multi-view clustering technique introduced by Grozavu et al. (2022) \cite{b2} because it allows for the comparison of data that are equivalent but defined by distinct factors, and we have revised it in order to apply it to the multi-modal context. All of the distributed views in this case share the same units but are described differently. Here, all NMF factorizations will share the same number of centroid vectors.\par
As stated earlier, let $X = (x_1, x_2,\cdots,x_N) \in \mathbb{R}_+^{M\times N}$ be a data set of $M$ features and $N$ objects containing non-negative values. In the case of multi-modal multi-view framework, we assume that we have a finite number of modalities $p \in \mathbb{N}$, and each modality has a finite number of views $v \in \mathbb{N}$. Locally, we apply a traditional NMF to each modality views. The local NMF expression can be rewritten as follows:
\begin{equation}\label{eq:nmf_colf}
\mathcal{L}_{(v_p)}(X^{(v_p)}, F^{(v_p)}, G^{(v_p)}) = \Vert X^{(v_p)} - F^{(v_p)}G^{(v_p)} \Vert^2
\end{equation}
where, the subscript $v_p$ denotes both the modality and view dependency. $F^{(v_p)} = (f_1^{(v_p)}, f_2^{(v_p)},\cdots,f_k^{(v_p)}) \in \mathbb{R}_+^{M\times K}$ is the cluster centroids matrix and $G^{(v_p)} = (g_1^{(v_p)}, g_2^{(v_p)},\cdots,g_N^{(v_p)}) \in \mathbb{R}_+^{K\times N^{(v_p)}}$ indicates the data partition matrix.

\subsection{Multi-view collaboration term}
Adding extracted information from different views $v' \neq v$ to a view $v$ is a popular collaborative approach \cite{b18} \cite{b19}. In their work \cite{b2}, Grozavu et al. presented a multi-view collaboration technique that minimizes the distance between a data point and its corresponding prototypes of local NMF views $v' \neq v$ to incorporate the information from view $v'$. In order to achieve this information transfer, they introduced the euclidean distances matrix $D^{(v)}$ of each data point of $X^{(v)}$ and the set of centroids $F^{(v)}$, such that $D_{kn}^{(v)} = \lVert x_n^{(v)} - k^{(v)}\rVert$.\par
However, this setting can’t be applied in our multi-modal context, since the euclidean distance is not suited for image similarities. As described in \cite{b20}, the euclidean distance is highly sensitive to even small image deformations. Since the traditional euclidean distance is a summation of the pixel-wise intensity differences, even minor deformations may produce large euclidean distances. Instead, when dealing with multi-modalities Hu et al. (2019) \cite{b21} suggest to use the inner product between each data point and the set of centroids. Taking this into account, we modify the distance matrix $D^{(v)}$ presented earlier by using the inner product instead of the euclidean distance, such that $d_{kn}^{v_p} = \langle x_n^{(v_p)},k^{(v_p)} \rangle$.\par
As a result, the pairwise collaborative term $\mathcal{C}(v_p, v'_p)$ between the $v^{th}$ and $v'^{th}$ NMFs is defined as follows:
\begin{equation}\label{eq:mv_col}
\mathcal{C}_{v_p,v'_p}(F^{(v_p)}, G^{(v_p)}) = \Vert (G^{(v_p)} - G^{(v'_p)}) \circ D^{(v_p)} \Vert_F^2
\end{equation}
Notice that $v'_p$ denotes another view of the same modality. The collaborative term $C(v_p,v'_p)$ is equivalent to the weighted sum of the inner product between the data point $x_n^{(v_p)}$ and all the centroids in $F^{(v_p)}$, with $G^{(v_p)} - G^{(v'_p)}$ representing the weight.\par
In (\ref{eq:mv_col}), when $G^{(v_p)}$ and $G^{(v'_p)}$ agree, the collaborative term equals zero and we consider only the $v^{th}$ local NMF

\subsection{Multi-modal collaboration term}
In our multi-modal multi-view context, we also want to include the information extracted from the views of the other modalities $p' \neq p$. With this additional knowledge transfer, the NMF algorithm not only include the information from local views $v'_p \neq v_p$ but also the information from distant views $v_{p'}$ (i.e. views of other modalities).\par
We define the multi-modal collaborative term as follows: 
\begin{equation}\label{eq:mm_col}
\mathcal{O}_{v_p,v_{p'}}(F^{(v_p)}, G^{(v_p)}) = \Vert F^{(v_p)}(G^{(v_p)} - G^{(v_{p'})})\Vert_F^2
\end{equation}
Having two data partition matrices of different modalities $G^{(v_p)}$ and $G^{(v_{p'})}$, our objective is to minimize the multi-modal collaborative term. Notice that $\mathcal{O}(v_p,v_{p'})$ is equal to zero if $G^{(v_p)}$=$G^{(v_{p'})}$.
\medbreak
Hence, the set of matrix partitions $G^{(v_p)}$ and the set of centroids $F^{(v_p)}$ are estimated iteratively and alternatively by minimizing the following objective function:
\begin{equation}\label{eq:obj_fct}
\resizebox{\hsize}{!}{$\mathcal{J}(F,G) = \sum_{p=1}^P \left(\sum_{v_p=1}^{V}(\mathcal{L}_v(F^{(v_p)}, G^{(v_p)}) + \mathcal{G}(v_p,v_p')\right) + \mathcal{H}(v_p,v_{p'})$}
\end{equation}
where
\begin{equation}\label{eq:g}
\mathcal{G}(v_p,v_p') = \sum_{v_p' \neq v_p}\beta_{v_p,v_p'}\cdot \mathcal{C}(v_p, v_p')
\end{equation}
and 
\begin{equation}\label{eq:h}
    \mathcal{H}(v_p,v_{p'}) = \sum_{v_p \neq v_{p'}} \gamma_{v_p,v_{p'}}\mathcal{O}(v_p,v_{p'})
\end{equation}
Here, $\mathcal{L}_v$ is the $v^{th}$ local NMF expression introduced in (\ref{eq:nmf_colf}). $\beta_{v_p,v_p'}$ and $\gamma_{v_p,v_{p'}}$ are the degrees of the multi-view and multi-modal collaborations respectively, with respect to the constraints $\sum_{v'_p\neq v_p}\beta_{v_p,v_p'} = 1$ and $\sum_{v_{p'}\neq v_p}\gamma_{v_p,v_{p'}} = 1$

\section{Optimization}
\subsection{Algorithm Derivation}
Recall that the described cost function in (\ref{eq:obj_fct}) is differentiable and its derivative exists at each point in its domain. As a result, there is always a minimum, which can be found using nonlinear programming.\par
To minimize the aforementioned cost function \eqref{eq:obj_fct}, we use the gradient descent technique. For $\Theta \in \{F^{(v_p)}, G^{(v_p)}\}$ $st.\hspace{0.2cm} \Theta \geq 0$, the update formula of the cost function (\ref{eq:obj_fct}) is:
\begin{equation}\label{eq:update_form}
\Theta = \Theta - \eta_{\Theta} \circ \nabla_\Theta (\mathcal{J}(F,G))
\end{equation}
Due to the presence of the subtraction operator in (\ref{eq:update_form}), the non-negativity condition is violated. To adress this issue, we consider Lee and Seung's strategy (2001) \cite{b22} by using an adaptive learning rate for the cost function $\mathcal{J}$ and the parameter $\Theta \in \{F^{(v_p)}, G^{(v_p)}\}$:
\begin{equation}\label{eq:adapt_lr}
\eta_\Theta = \frac{\Theta}{[\nabla_\Theta \mathcal{J}]_+}
\end{equation}
The update rule of the partition matrix and centroid matrix is written as follows:
\begin{equation}\label{eq:multi_update}
\Theta = \Theta \circ \frac{[\nabla_{\Theta}\mathcal{J}]_-}{[\nabla_{\Theta}\mathcal{J}]_+}
\end{equation}
Such that $\circ$ and the fraction line represent the element-wise multiplication and division respectively. Notice that in (\ref{eq:multi_update}), the negative terms of the gradient are in the numerator, while the denominator contains the positive terms.\par

\subsection{Optimized Weights for the Collaborative terms}
Here, we examine how optimizing the degrees of collaboration $\beta_{v_p,v_p'}$ and $\gamma_{v_p,v_{p'}}$, introduced in \eqref{eq:g} and \eqref{eq:h}, can produce the optimal solution for the cost function and reduce the risk of negative collaboration.\par
Since $\beta_{v_p,v_p'} \geq 0$, we consider the collaboration weight $\beta_{v_p,v_p'} = \tau^2_{v_p,v_p'}$. Our objective is to find the positive weights $\tau^2_{v_p,v_p'}$ that will determine the collaborative term's strength. Using the condition $\forall v_p$, $\sum_{v'_p \neq v_p}^{V} \tau^2_{v_p,v_p'} = 1$ along with the Karush-Kuhn-Tucker (KKT) conditions \cite{b23}, the results of the optimization are presented in (\ref{eq:beta}):
\begin{equation}\label{eq:beta}
\beta_{v_p,v_p'} = \frac{|\mathcal{C}_{v_p,v_p'}|^2}{\left(\sum_{v'_p \neq v_p} |\mathcal{C}_{v_p,v_p'}|^2\right)}
\end{equation}
Similarly, the optimized multi-modal collaborative term is written as:
\begin{equation}\label{eq:gamma}
\gamma_{v_p,v_{p'}} = \frac{|\mathcal{O}_{v_p,v_{p'}}|^2}{\left(\sum_{v_{p'} \neq v_p} |\mathcal{O}_{v_p,v_{p'}}|^2\right)}
\end{equation}
We propose an interpretation to these results: in the context of multi-modal multi-view collaboration, overall results should improve if individual algorithms give more weight to algorithms with the same results as local solutions(higher weight $\tau$ value for a specific NMF model). 

\begin{algorithm}[H]
\SetAlgoLined
\label{alg:algoG}
\vspace{0.05cm}
\caption{Multi-modal Multi-view  NMF}

\vspace{0.05cm}
\textbf{Convex-initialization:} Randomly set the cluster prototypes, $v$ number of views and $p$ number of modalities. \\
\textbf{For all realizations} \\
\textbf{Local phase:} \\
      \ForAll{views $v$ of a modality $p$} {
      	Optimize the NMF cost function   \eqref{eq:nmf_colf}.
      } 
\textbf{Multi-modal Multi-view Collaboration phase:}\\
                 \textbf{Compute the optimized $\beta_{v_p,v_p'}$ with \eqref{eq:beta}} \\
                 \textbf{Compute the optimized $\gamma_{v_p,v_{p'}}$ with \eqref{eq:gamma}} \\
		 \ForAll{views $v$ of all modalities $p$} {
		  Estimate the partitions matrix of all views  \eqref{eq:multi_update}. \\ 	 
		  Estimate the centroids matrix of all views  \eqref{eq:multi_update}. 
 	 }
	 	  
\end{algorithm}

\section{Experiments}
In this section, we assess the performances of our proposed collaborative strategy on two multi-modal data sets:  Multimodal Corpus of Sentiment Intensity (MOSI) \cite{b24} and NUS-WIDE \cite{b25}. Further details on the data sets are given in order to illustrate the premise of the presented approach. Since we have access to these data set labels, the performance of the multi-modal multi-view NMF clustering is evaluated using two standard metrics: the silhouette index and purity.

\subsection{Purity Evaluation Procedure}
Purity is a metric that measures the extent to which clusters contain a single class. Let $L = \{l_1, l_2, \cdots l_n\}, n \in \mathbb{N}$ and $K = \{k_1, k_2, \cdots k_m\}, m \in \mathbb{N}$ be the known data labels and centroids respectively. The purity score of a clustering is defined as:
\begin{equation}\label{eq:purity}
    purity = \sum_{m=1}^{|K|} 
\dfrac{\max_{i=1}^{|L|}|k_{im}|}{|k_{m}|}
\end{equation}
where $|k_{m}|$ denotes the total number of observations associated with the cluster $k_m$, and $|k_{im}|$ denotes the amount of data of class $l_i$ related to the cluster $k_m$.\par
The purity of the clustering result is equal to the expected purity of all clusters. A High purity score indicates a good clustering process.

\subsection{Silhouette Evaluation Procedure}
The silhouette index is the average silhouette coefficient over each data sample. It is computed using the following formula:
\begin{equation}\label{eq:silhouette}
    silhouette = \frac{(b-a)}{\max(a,b)} 
\end{equation}
where $a$ is the mean distance between instances of the same cluster (i.e. the mean intra-cluster distance), and $b$ is the mean distance to the instances of the successive closest cluster (i.e. mean nearest-cluster distance).\par
The silhouette coefficient is defined in the interval $[-1,1]$; a value close to $1$ indicates that the instance is inside its own cluster and distant from other clusters, a value close to 0 indicates that it is near a cluster boundary, and a value close to $-1$ indicates that the instance may have been mistakenly assigned to a different cluster.
\subsection{Data Set Descriptions}
\begin{itemize}
    \item \textbf{NUS-WIDE -} contains 269,648 images and their associated 5,018 unique tags from Flickr. A ground truth of 81 classes is provided, consisting of events, programs, animals, objects, people. A semi-automatic process is used to create the ground truth and human labelers assess the relevance of the image classes. Six low-level image features are given: color histogram, color correlogram, edge direction histogram, wavelet texture, block-wise color moments and a bag of visual words on SIFT descriptions.\par
    We extracted two subsets (NUS-2B, NUS-CDF) that we used for our experimentation (see Tab. \ref{tab:nus_subset}). Experiments are performed using both modalities (i.e. image and text). For the image modality, we used the edge direction histogram and the wavelet texture.
    \item \textbf{MOSI -} contains 2199 opinion video clips. Each clip has a sentiment annotation in the interval $[-3, 3]$.For each opinion video clip, the audio file and transcriptions are provided. The data set is meticulously annotated with labels for sentiment intensity and subjectivity.\par
    In our experimentation, we only used the text and audio modalities. We extracted two views (low-level features) from the audio modality: the raw audio signal (Raw) and the Mel-scale spectrogram (MEL), as suggested in \cite{b26}. As for the text modality, we used the BERT \cite{b27} and Word2Vec (W2V) \cite{b28} views. To illustrate the process of multi-modal multi-view collaboration, we introduce a Gaussian noise with a mean of zero and a standard deviation of one to the Word2Vec view. Finally, in order to compute the purity score, we transformed the data set into binary classification by assigning the label "\textit{positive}" to the sentiments in the interval $(0, 3]$ and the label "\textit{negative}" to the sentiments in the interval $[-3, 0]$
\end{itemize}
\begin{table}[b]
\caption{Sub-datasets of NUS-WIDE}
\begin{center}
\begin{tabular}{|l|l|l|}
\hline
\textbf{Data sets} & \textbf{Classes}    & \textbf{Images}  \\ \hline
NUS-CDF & \begin{tabular}[c]{@{}l@{}}Cat\\ Dog\\ Fish\end{tabular} & \begin{tabular}[c]{@{}l@{}}1425\\ 1486\\ 1019\end{tabular} \\ \hline
NUS-2B  & \begin{tabular}[c]{@{}l@{}}Bird\\ Boat\end{tabular}      & \begin{tabular}[c]{@{}l@{}}2224\\ 2477\end{tabular}        \\ \hline
NUS-WIDE  & 81 classes & 269 648 \\ \hline
\end{tabular}
\label{tab:nus_subset}
\end{center}
\end{table}

\subsection{Illustration of the proposed solution on the NUS-2B subset}
As stated previously, we will use the case of a collaboration between two views of the same modality (image) and a view of the other modality (text) to simplify the interpretation of the collaboration principle.\par
To allow collaboration between different views and modalities, the structures of all local clustering results must be similar (i.e. same dimensions). To ensure this condition, we applied PCA (Principal Component Analysis) on all modalities views.
\begin{figure}[b]
    \centering
    \includegraphics[width=0.7\columnwidth]{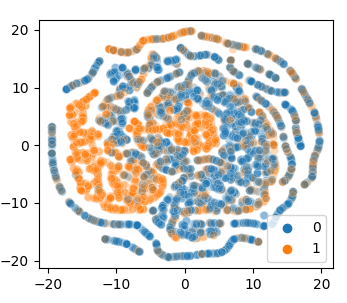}
    \caption{Wavelet texture view representation prior to collaboration}
    \label{fig:wt_before}
\end{figure}

Fig. \ref{fig:wt_before} represents a projection of the wavelet texture view in a two-dimensional space using T-SNE (T-Distributed Stochastic Neighbor Embedding), using the ground truth provided. The associated cluster for each set of data is displayed with a specific color.\par
Using NMF prior to the multi-modal multi-view collaboration, the purity scores achieved on the image views (wavelet texture and edge direction histogram) and the text view are $52.7\%$, $71.6\%$ and $85.6\%$ respectively.\par
We then applied the second phase of the proposed method (the multi-modal multi-view collaboration) to share the clustering information throughout all NMF clustering results.\par
\begin{figure}[t]
    \centering
    \includegraphics[width=0.7\columnwidth]{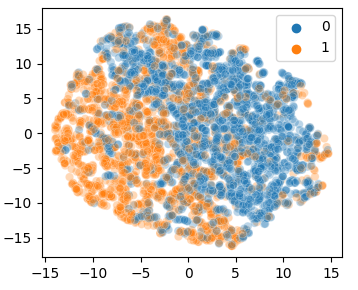}
    \caption{Wavelet texture view representation after the collaboration}
    \label{fig:wt_after}
\end{figure}

Following the collaboration of the edge direction histogram and text views with the wavelet texture view, the purity score of the latter rose to $66.4\%$. Fig. \ref{fig:wt_after} shows the result of the muti-modal multi-view collaboration on the wavelet texture image view.
Furthermore, we computed the Silhouette index to evaluate the resulting clustering structure after the collaboration. The Silhouette index increased from 0.32 to 0.38. Tab. \ref{tab:nus2b_sum} summarizes these experiments.\par
In another experiment, we analyzed the impact of the horizontal collaboration of views with lower purity score on a view with a higher score. To do so, we introduced a Gaussian noise to the text view to reduce its clustering quality, which became $61.85\%$. Next, by collaborating the noisy text and the wavelet texture views with the edge histogram texture view, the purity score of the latter decreased from $71.6\%$ to $63.08\%$.\par
We notice that the collaboration between a view with a low purity score and views and modalities with higher purity scores enhances the quality of the initial view. Whereas, a collaboration between a view with a higher purity score and views and modalities with lower purity score diminishes the quality of the initial view.\par
These findings indicate that while the multi-modal multi-view collaboration increases or decreases the purity score based on the clustering quality of distant collaborators, it has a little impact on the Silhouette index, as the collaboration only takes in consideration the distant partitions without altering the local structure of the view.\par
\begin{table}[t]
\caption{Results of the horizontal collaboration method on NUS-2B}
\centering
\begin{tblr}{
  width = \linewidth,
  colspec = {Q[156]Q[460]Q[113]Q[187]},
  cell{1}{1} = {r=2}{},
  cell{1}{2} = {r=2}{},
  cell{1}{3} = {c=2}{0.3\linewidth},
  cell{3}{1} = {r=6}{},
  vlines,
  hline{1,3,9} = {-}{},
  hline{2} = {3-4}{},
  hline{4-8} = {2-4}{},
}
Dataset & NMF                       & Metrics &            \\
        &                           & Purity     & Silhouette \\
NUS-2B  & NMF$_{edh}$                  & 71,6       & 0,34       \\
        & NMF$_{wt}$                   & 52,7       & 0,32       \\
        & NMF$_{text}$                 & 85,6       & 0,37       \\
        & NMF$_{noisyText}$            & 61,85      & 0,33       \\
        & NMF$_{edh,text->wt}$        & 66,4       & 0,38       \\
        & NMF$_{wt,noisyText->edh}$ & 63,08      & 0,3        
\end{tblr}
\label{tab:nus2b_sum}
\end{table}

\subsection{Comparison with other technique}
To illustrate the usefulness of the multi-modal multi-view collaboration approach presented, we compare it with the multi-view clustering technique proposed in \cite{b2}. The comparison is conducted on the NUS-CDF subset.\par
Tab. \ref{tab:tech_comp} gives the purity score and Silhouette index of each local NMF clustering and horizontal collaboration algorithm. Regarding the Multi-view NMF clustering technique, the clustering quality of the edge direction histogram view decreased from $39.3\%$ to $38.32\%$, after the collaboration, due to the local knowledge transfer of the wavelet texture view (lower purity of $37.98\%$). Whereas, using our method, the edge direction histogram view's clustering quality increased, after the collaboration, from $39.3\%$ to $57.04\%$ as a result of the local knowledge transfer of both the text modality and the intra-modality view (wavelet texture).\par
This comparison shows the importance of including the information from other modalities during the collaboration.
\begin{table}[t]
\caption{Comparison of Approaches on NUS-CDF subset}
\centering
\begin{tblr}{
  width = \linewidth,
  colspec = {Q[275]Q[346]Q[113]Q[185]},
  cell{1}{1} = {c=2}{0.621\linewidth},
  cell{2}{1} = {r=3}{},
  vlines,
  hline{1-2,5-7} = {-}{},
  hline{3-4} = {2-4}{},
}
Algorithms     &                    & Purity & Silhouette \\
NMF            & NMF$_{edh}$           & 39,3   & 0,31       \\
               & NMF$_{wt}$            & 37,98  & 0,34       \\
               & NMF$_{text}$          & 94,75  & 0,59       \\
Multi-view NMF & NMF$_{wt->edh}$        & 38,32  & 0,32       \\
Our approach   & NMF$_{wt,text->edh}$ & 57,04  & 0,36       
\end{tblr}
\label{tab:tech_comp}
\end{table}

\subsection{Validation using additional data sets}
In this part, we applied our solution to the MOSI data set and computed the clustering purity score before and after the collaboration.\par
In Tab. \ref{tab:mosi_sum}, notice that the purity score increases when the majority of distant collaborators have a strong segmentation. Similarly, we can see that the collaboration has little impact on the Silhouette index since the data set structure remains unchanged.

\begin{table}[h]
\caption{Results of the horizontal collaboration method on MOSI.}
\centering
\begin{tblr}{
  width = \linewidth,
  colspec = {Q[179]Q[387]Q[131]Q[212]},
  cell{1}{1} = {r=2}{},
  cell{1}{2} = {r=2}{},
  cell{1}{3} = {c=2}{0.343\linewidth},
  cell{3}{1} = {r=5}{},
  vlines,
  hline{1,3,8} = {-}{},
  hline{2} = {3-4}{},
  hline{4-7} = {2-4}{},
}
Dataset & NMF              & Metrics &            \\
        &                  & Purity     & Silhouette \\
MOSI  & NMF$_{BERT}$        & 55,34      & 0,39       \\
        & NMF$_{W2V}$         & 46,2       & 0,31       \\
        & NMF$_{Raw}$         & 51,79      & 0,36       \\
        & NMF$_{MEL}$         & 52,88      & 0,35       \\
        & NMF$_{B, R, M->W}$ & 53,11      & 0,33       
\end{tblr}
\label{tab:mosi_sum}
\end{table}

\section{Conclusion}
In this study, we presented a novel method for multi-modal multi-view horizontal collaboration by transferring knowledge between various local Non-negative Matrix Factorizations. Through this collaboration, various NMFs can interact and reveal the inherent patterns and structures in data sets.\par
We presented our proposed technique, which is well-suited for collaboration between views of various modalities that represent the same objects but with different attributes.\par
The experimental findings show that the proposed method, which has been validated against a variety of data sets, produces better results than the multi-view NMF clustering.\par
As part of our future work, we plan to implement an ensemble technique to find a single consensus partition among all the local NMFs after the collaboration. We also plan to analyze the impact of different modalities on the results of the collaboration by introducing a weight factor for each modality.



\end{document}